\documentclass[letterpaper, 10 pt, conference]{ieeeconf}  
\IEEEoverridecommandlockouts                              

\usepackage{color}
\usepackage[utf8]{inputenc}
\usepackage{amsmath,amssymb}
\usepackage{mathtools}
\usepackage{amsthm}
\usepackage{graphicx}
\usepackage{float}
\usepackage{wrapfig}
\usepackage{algorithmicx,algpseudocode}
\usepackage{algorithm}
\usepackage{amsfonts,dsfont}
\usepackage{mathrsfs}  
\usepackage{booktabs}

\renewcommand{\phi}{\varphi}
\renewcommand{\epsilon}{\varepsilon}

\usepackage{hybridsystem}

\theoremstyle{plain}

\theoremstyle{remark}
\newtheorem{rem}{\textbf{Remark}}

\newcommand{\drs}{12} 

\title{\LARGE \bf 
Data-driven Characterization of Human Interaction for Model-based Control of Powered Prostheses}
\author{Rachel Gehlhar, Yuxiao Chen, and Aaron D. Ames
\thanks{*This material is based upon work supported by the National Science Foundation Graduate Research Fellowship under Grant No. DGE‐1745301 and NSF NRI Grant No. 1724464.}
\thanks{R. Gehlhar, Y. Chen, and A. Ames are with the Department of Mechanical and Civil Engineering, California Institute of Technology, Pasadena, CA 91125 USA. Emails:
{\tt\small $\{$rgehlhar, chenyx, ames$\}$@caltech.edu}}
}

\begin{document}

\maketitle

\begin{abstract}
This paper proposes a data-driven method for powered prosthesis control that achieves stable walking without the need for additional sensors on the human. The key idea is to extract the nominal gait and the human interaction information from motion capture data, and reconstruct the walking behavior with a dynamic model of the human-prosthesis system. The walking behavior of a human wearing a powered prosthesis is obtained through motion capture, which yields the limb and joint trajectories. Then a nominal trajectory is obtained by solving a gait optimization problem designed to reconstruct the walking behavior observed by motion capture. Moreover, the interaction force profiles between the human and the prosthesis are recovered by simulating the model following the recorded gaits, which are then used to construct a force tube that covers all the interaction force profiles. Finally, a robust Control Lyapunov Function (CLF) Quadratic Programming (QP) controller is designed to guarantee the convergence to the nominal trajectory under all possible interaction forces within the tube. Simulation results show this controller's improved tracking performance with a perturbed force profile compared to other control methods with less model information.

\end{abstract}


\section{INTRODUCTION}
Commercially available prosthetic legs remain largely limited to passive devices which increase an amputee's metabolic cost and their amputated side's hip power and torque \cite{winter1991biomechanics}.
Powered prostheses lend the benefit of providing net power to the user and enabling a walking gait more representative of a healthy biomechanical gait \cite{lawson2012preliminary}.
A large subset of existing research on powered prostheses focuses on the use of impedance control methods \cite{PowAnkleFootStair, DesignControlTransProsth, VirtConsCtrlProst}. The downsides of this method is that it requires extensive tuning and is highly heuristic. To address this heuristic nature, researchers have developed trajectory tracking methods for prostheses inspired by bipedal robotics \cite{VirtConsCtrlProst, Stein1987StancePC, CtrlWearRobot}.

Powered prostheses present an interesting control problem compared to walking robots in that the behavior of part of the system is unknown: the human. To address this, researchers have examined phase variables \cite{VirtConsCtrlProst, zhao2016multi} to properly modulate the prosthesis trajectory in response to the human, but the trajectory tracking methods do not account for the human dynamics. 
While feedback linearization and CLFs can enforce the trajectories on walking robots \cite{ames2012control, ames2013human}, these methods cannot be applied to prostheses in the same way because they require full model information. 
For example, in \cite{zhao2017preliminary} a method was developed to apply CLF-QP to a prosthesis, but in a model independent fashion. Model-dependence is desired in a prosthesis controller to improve tracking performance and robustness to perturbations. Recent work \cite{azimi2017robust} incorporated some model dependence into robust prosthesis controllers but did not account for the interaction force between the human and prosthesis.

Accounting for the interaction force in a model-dependent prosthesis controller is crucial for the stance phase for two reasons. One, during stance the human exerts a large force on the prosthetic as the prosthetic supports the human's weight and motion, hence the force is a critical component of the prosthesis dynamics. Two, the human relies on the prosthetic for support and balance, making the stability of the prosthesis vital for the human's safety during this phase. 
Prosthesis controllers were developed in \cite{HybInvStabFeedLin} and \cite{gehlhar2019control} that incorporated this interaction force in feedback linearization. However, while these methods worked in simulation, they pose implementation problems due to the drawbacks of a force sensor and lack of robustness of feedback linearization. Force sensors for these applications are expensive, noisy, and not robust to the multi-directional force and torque impacts present in walking. This motivates our goal to develop a model dependent prosthesis control method without requiring a force sensor.

\begin{figure} [t!] 
\label{fig:Ampro3} 
\centering
\includegraphics[width=1\columnwidth]{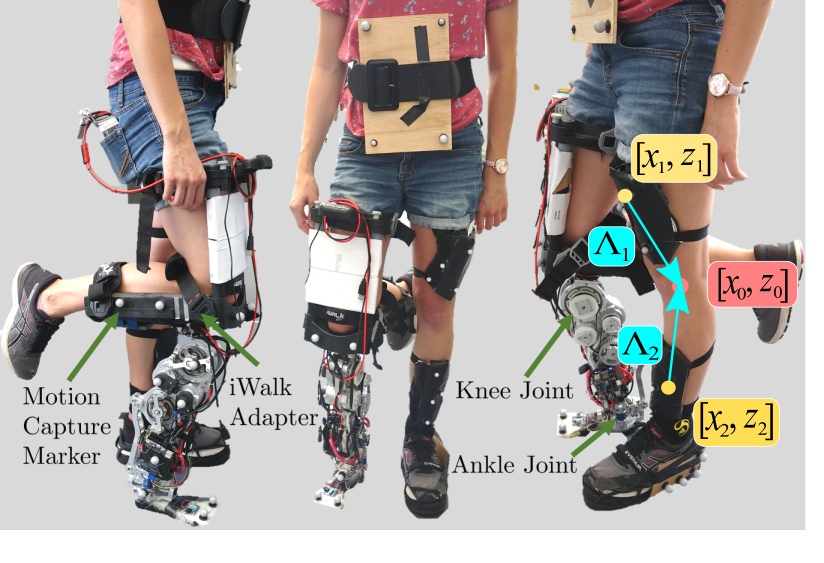}
{\caption{Powered prosthesis AMPRO3 attached to iWalk adapter worn by human with Optitrack motion capture markers. (Right) Joint location determination based on markers.}}
\vspace{-0.6cm}
\end{figure}

In order to characterize the reaction force between a human and prosthesis without a force sensor, human walking analysis is needed.
Motion capture has often been used to understand human walking behavior \cite{dasgupta1999making, ames2014human}. 
The authors in \cite{yang2015mechanical} and \cite{zhao2017first} used motion capture to develop a reference trajectory for a powered prosthesis, yet the reference is taken from normal human walking. 
In \cite{wehner2009internal}, the internal force of human lower-limbs was analyzed with a musculoskeletal model, yet the model is only applicable to a human body, not a human body in connection with a prosthesis. This paper extends upon these works by \textit{using motion capture to evaluate the interaction force} between a human user and prosthesis device and develop a stable walking trajectory for the prosthesis. 

The main contributions of this paper are 
\begin{itemize}
    \item developing a procedure that identifies the range of interaction forces on the prosthesis from motion capture without a force sensor
    \item constructing a robust CLF-QP controller that renders the prosthesis stable, even with force disturbances, to a walking trajectory similar to that in motion capture.
\end{itemize}
We obtain joint trajectory data using an Optitrack motion capture system, and then calculate the interaction forces by simulating the human-prosthesis system following the trajectories. Through optimization we match the trajectories to obtain a stable nominal walking gait that satisfies the dynamic equations to simulate continuous walking. For this we follow a method similar to \cite{ames2014human} but use asymmetric human-prosthesis data with a human-prosthesis model instead of symmetric human data. Then, to capture the nondeterministic nature the of a human user, a force tube is constructed that covers all interaction forces associated with the recorded steps. Finally, we control the prosthesis in simulation with a robust CLF-QP controller that guarantees convergence to the nominal gait for any possible interaction force within the tube. Simulation results show the improved tracking performance of this model-dependent method with respect to perturbations. 

The structure of this paper is depicted in the flow diagram in Fig. 2 and is outlined here.
Section \ref{sec:MotionCapture} explains the method used to obtain human-prosthesis walking data through motion capture and how the data was processed to obtain joint angle trajectories. Section \ref{sec:human-prosth-model} describes the hybrid model for the combined human-prosthesis system. Section \ref{sec:Simulate} covers the construction of outputs for the human-prosthesis system and how these are used in gait design to both develop a stable walking trajectory that matches the motion capture data and playback the data to obtain the interaction force profiles. Further, this section covers the development of the tube of interaction forces. Section \ref{sec:RobustCLF} outlines the construction of our robust CLF-QP controller and presents the simulation results with this controller and our developed walking trajectory. Finally Section \ref{sec:Conclusion} concludes the paper and contains a brief description of future work.

\begin{figure}
\centering
\includegraphics[width=1\columnwidth]{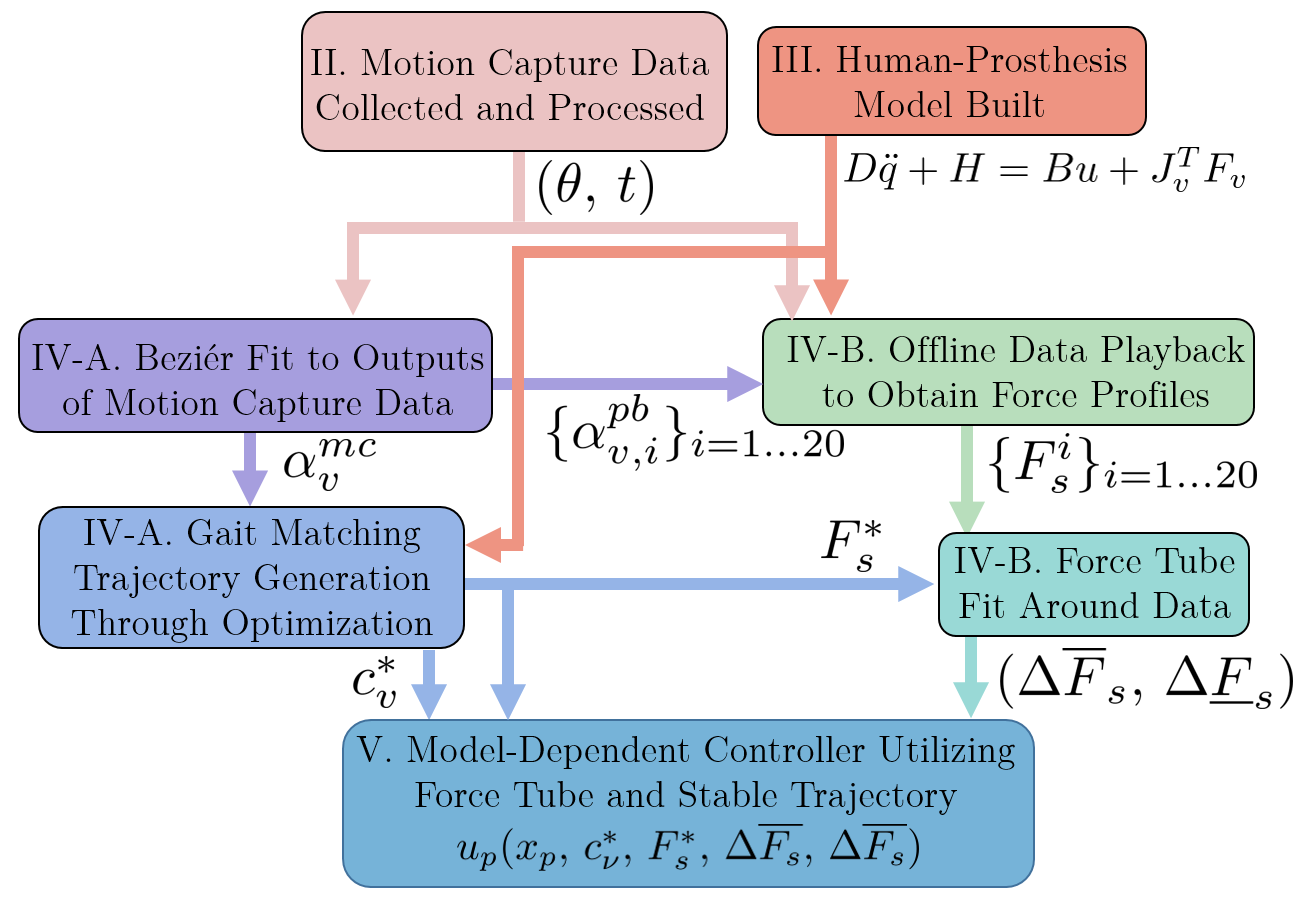}
{\caption{Flow diagram (with section numbers of paper) depicting steps in method to go from motion capture data to a model-dependent controller.}}
\vspace{-0.6cm}
\end{figure}
 
\section{MOTION CAPTURE AND DATA PROCESSING} \label{sec:MotionCapture}
In the data collection phase, a human user wears the powered transfemoral prosthesis, AMPRO3, through the use of an iWalk adapter on her right leg, as shown in Fig. 1.
AMPRO3 has 3DOF, active knee pitch, active ankle pitch, and passive ankle roll \cite{zhao2017preliminary}. The knee and ankle are controlled in real-time by the most model-dependent prosthesis controller currently available, a robust-passive controller that tracks a trajectory determined offline \cite{azimi2017robust}.

The behavior of the human user with prosthetic leg is captured by the motion capture system Optitrack installed in the Center for Autonomous Systems and Technologies (CAST) at California Institute of Technology (Caltech). It uses multiple cameras (up to 40) to locate markers fixed on an object and subsequently locate the object. We place markers on every limb of the human user and AMPRO3, registering each set of markers on the same limb as a rigid body in the tracking system. Optitrack gives the position and rotation of the limbs.

\subsection{Processing Motion Capture Data} \label{ssec:ProcessMCData}
Since the prosthesis only actuates in the sagittal plane, we project the motion data to this plane, treating the data as 2D walking data. For this section, let $x,z$ denote the longitudinal and vertical coordinates of the markers and $\theta$ denote the pitch angle of the rigid bodies. Optitrack gives the global coordinates of the markers along with the position and orientation of the rigid bodies when it recognizes the rigid bodies. From this we compute the limb pitch angles, joint angles, and joint positions.

\newsec{Compute Pitch Angles.}
    When the system recognizes the rigid body, it directly provides the pitch angle.
    When the tracking is lost, we use two markers on the same limb to compute the pitch angle as $\theta  = \arctan(({x_2} - {x_1})/({z_2} - {z_1}))$,
    where $[x_i,z_i],i\in\left\{1,2\right\}$ are the $x$ and $z$ coordinates of the two markers. The difference between limb angles gives the joint angle, providing the joint angle trajectories for walking.
    
    \newsec{Locate Joint Positions.}
    Since markers are not on joint rotation centers, we use a convex optimization approach to determine the joint position relative to markers to obtain the global joint positions from the marker positions.
    For each joint, we find the two limbs connected to the joint, and select one marker on each joint, denoted as $[x_1,z_1]$, $[x_2,z_2]$, as shown in the right portion of Fig. 1. We signify the joint coordinate with $[x_0,z_0]$ and the previously determined limb pitch angles as $\theta_1$ and $\theta_2$.
    Let $\Lambda_1$ and $\Lambda_2$ denote the vectors from the two markers to the joint, then the joint position can be computed from two directions:
    \begin{equation}\label{eq:joint_position}
        [x_0,z_0]^\intercal=[x_1,z_1]^\intercal+R(\theta_1)\Lambda_1=[x_2,z_2]^\intercal+R(\theta_2)\Lambda_2,
    \end{equation}
    where $R(\theta)$
    is the 2D rotation matrix. 

  The following optimization solves for $\Lambda_1$ and $\Lambda_2$ by minimizing the discrepancy between the two equivalent computations: 
\begin{equation*}
\mathop {\min }\limits_{{\Delta _1},{\Delta _2}\in\mathbb{R}^2} \sum\nolimits_{t}{\left\| \begin{array}{l}{[
{{x_1}(t)},
{{z_1}(t)}]^\intercal + R({\theta _1}(t)){\Lambda _1} -}\\ 
{[
{{x_2}(t)},{{z_2}(t)}]^\intercal - R({\theta _2}(t)){\Lambda _2}}\end{array} \right\|_2},
\end{equation*}
with which we compute the joint location by taking the average of the two expressions in \eqref{eq:joint_position}. The joint position information is potentially useful for the computation of limb lengths, identifying different phases of the walking data, and computing outputs that depend on joint positions.

With the procedure presented in this section, we are able to get the limb angle trajectories, joint position trajectories, and joint angle trajectories from the motion capture data. In total, 37 step cycles were collected and analyzed.

\section{HUMAN-PROSTHESIS MODEL} \label{sec:human-prosth-model}
To simulate the joint trajectories from motion capture data, we develop a model of the human-prosthesis system. 

\newsec{Model.}
The human-prosthesis system is modeled as a 2D bipedal robot by the methods of \cite{ModelsGrizzle} with the addition of a 3 DOF fixed joint at the human-prosthesis interface, for a total of 12 DOF, shown in Fig. 3. We consider 6 actuators, one at each leg joint.
The human limb lengths are based on human measurements. The human limb mass and COM are calculated with Plagenhoef's table of percentages \cite{HumanParam} and the subject’s total mass. The inertia of each limb is estimated with Erdmann's table of radiuses of gyration \cite{HumanInertia}. The human right thigh limb accounts for the iWalk, human’s bent calf, and human’s foot. We measured the iWalk’s mass and length and used these measurements to calculate the moment of inertia assuming simple geometry.
The prosthesis parameters are obtained from a CAD model of AMPRO3 \cite{zhao2017preliminary}, a powered transfemoral prosthesis, Fig. 1.

\begin{figure}
\centering
\includegraphics[width=1\columnwidth]{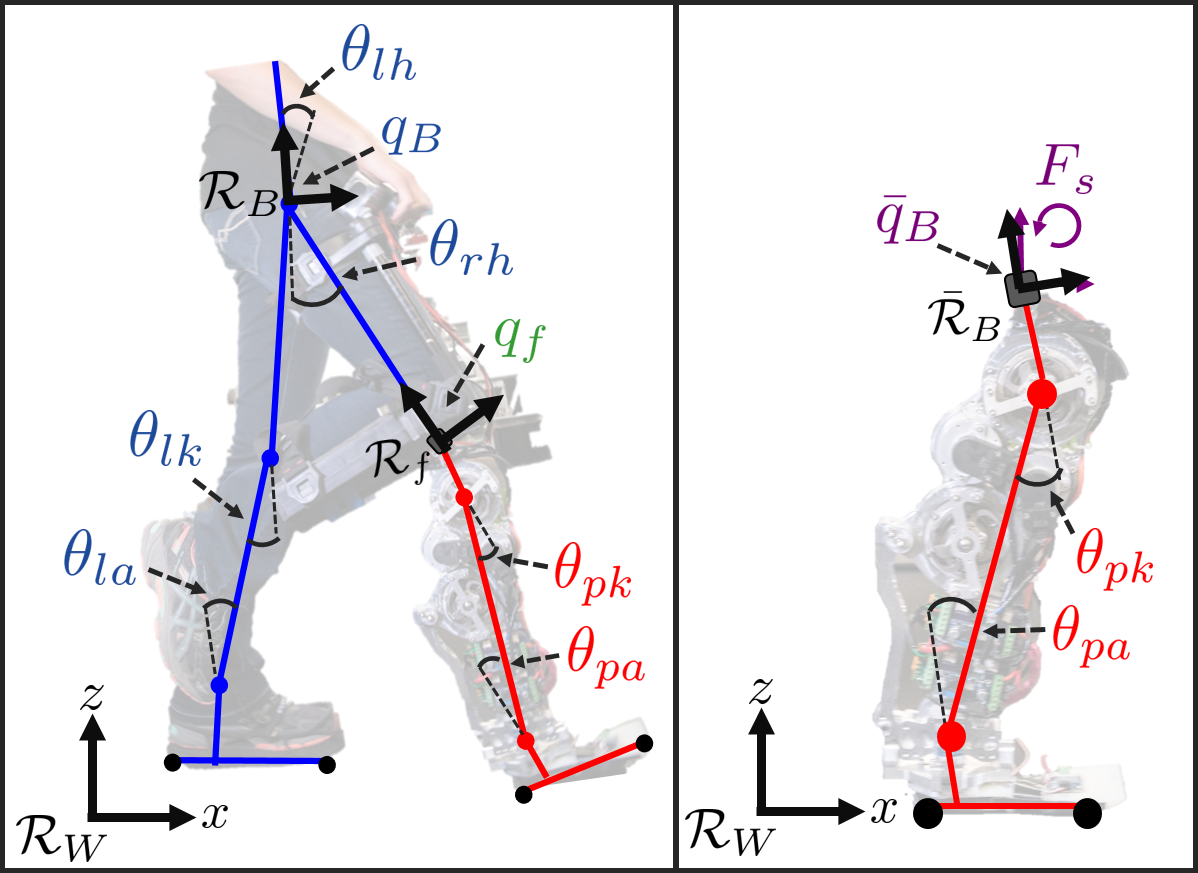}
{\caption{(Right) Robot model of human-prosthesis system labeled with generalized coordinates. (Left) Model of robotic prosthetic leg with external forcing $F_s$.}}
\vspace{-0.3cm}
\end{figure}

\newsec{Generalized Coordinates.}
We define the generalized coordinates for the model as $q = (q_h^\intercal,\, q_f^\intercal,\, q_p^\intercal)^\intercal$. Here, the coordinates of the human side are $q_h = (\q_B^\intercal,\, \theta_{lh},\, \theta_{lk},\, \theta_{la},\, \theta_{rh})^\intercal$, where the extended coordinates $\q_B \in SE(2)$ represent the position and rotation of the system's base frame $\mathcal{R}_B$ with respect to the world frame $\mathcal{R}_W$, and the remaining coordinates are the relative joint angles as defined in Fig 3. The coordinates of the fixed joint $q_f \in \mathbb{R}^3$ are the position and rotation of the fixed joint reference frame $\mathcal{R}_f$. The prosthetic coordinates are given by $q_p = (\theta_{pk},\, \theta_{pa})^\intercal$, for the knee and ankle, respectively. 

\newsec{Human-Prosthesis Dynamics.}
Because human walking contains both continuous and discrete dynamics, we model it as a \textit{multi-domain hybrid control system}, formally defined as a tuple \cite{ames2014human}: 
\begin{equation*}
    \HybridControlSystem = (\DirectedGraph,\, \Domain,\, \ControlInput,\, \Guard,\, \Delta,\, FG),
\end{equation*}
where $\DirectedGraph = (V,\, E)$ is a \textit{directed cycle}, with vertices $V = \{v_1 = ps,\, v_2 = pns\}$ and edges $E = \{e_1 = \{ps \rightarrow pns\}, e_2 = \{pns \rightarrow ps\}\}$. 
Here \textit{ps} stands for \textit{prosthesis stance} and \textit{pns} for \textit{prosthesis non-stance}. 
These are modeled as separate domains because of the asymmetry of the model.
(For the scope of this paper, we only consider these single support phases since we are most interested in controlling the prosthesis when it is the only support for the human.
We assume when the human has her own foot as support, she can balance herself more independently.)
Each domain $\Domain_{v}$, in the set of admissible domains defined by $\Domain = \{\Domain_{v}\}_{v \in V}$, contains two 3-DOF holonomic constraints, $h_v(q) \in \mathbb{R}^6$, one on the stance foot and the other on the fixed joint at the human-prosthesis interface. 
The set of admissible inputs is defined by $\ControlInput = \{\ControlInput_{v}\}_{v \in V}$. 
The transitions between the domains are a set of guards, $\Guard = \{\Guard_{e}\}_{e \in E}$, which in this case is when the non-stance foot hits the ground. This event causes an impact defined by $\Delta = \{\Delta_{e}\}_{e \in E}$. The set of control systems $FG = \{(f_v,\, g_v)\}_{v \in V}$ with $(f_v,\, g_v)$ defines the continuous dynamics $\dot{x} = f_v(x) + g_v(x)u_v$.

To obtain these continuous dynamics of the human-prosthesis system with $x = (q^\intercal, \dot{q}^\intercal)^T$,
we use the classical Euler-Lagrangian equation for robotic systems 
\cite{ModelsGrizzle}, \cite{MLS}:
\begin{equation} \label{eq:robotDyn}
D(q) \ddot{q} + H(q, \dot{q}) = B u + J_v^T(q) F_v(q, \dot{q}).
\end{equation}
\noindent Here $D(q) \in \mathbb{R}^{\drs \times \drs}$ is the inertial matrix. $H(q, \dot{q}) = C(q, \dot{q}) + G(q) \in \mathbb{R}^{\drs}$, a vector of centrifugal and Coriolis forces and a vector containing gravity forces, respectively. The actuation matrix $B \in \mathbb{R}^{\drs\times 6}$ contains the gear-reduction ratio of the actuated joints and is multiplied by the control inputs $u \in \mathbb{R}^{6}$. The wrenches $F_v(q, \dot{q}) \in \mathbb{R}^{6}$ enforce the $6$ holonomic constraints. 
The Jacobian matrix of the holonomic constraints $J_v(q) = \frac{\partial h_v}{\partial q} \in \mathbb{R}^{6 \times \drs}$ 
enforces the holonomic constraints by the following equation:
\begin{equation}\label{eq:holoConstr}
\dot{J}_v(q, \dot{q}) \dot{q} + J(q)_v \ddot{q} = 0.
\end{equation}
Solving \eqref{eq:robotDyn} and \eqref{eq:holoConstr} simultaneously yields the \textit{constrained dynamics}. 
These terms will now be referred to as $D,\, H,\, J_v,\, \text{and } F_v$, respectively, for notational simplicity.

\section{SIMULATING MOTION CAPTURE DATA} \label{sec:Simulate}
In order to reconstruct the motion capture in simulation, we track the joint trajectories obtained from motion capture with the human-prosthesis model built in Section \ref{sec:human-prosth-model}. Since the raw data does not satisfy the dynamics equations, we construct a stable reference trajectory close to the data in Section \ref{ssec:GaitDesign} to simulate continuous walking for controller testing. To estimate the range of interaction forces and moments seen by the prosthesis from the human for use in a controller, we simulate multiple continuous domains of the data in Section \ref{ssec:SocketForce}.

\subsection{Gait Design} \label{ssec:GaitDesign}
To design a stable walking gait, state-based outputs are defined to enable construction of a state-based controller for improved robustness \cite{westervelt2018feedback}. 

\newsec{State-based Outputs for Control.}
To modulate the outputs, a monotonic phase variable $\tau(q)$ is developed with the linearized hip position relative to the ankle:
\begin{equation*}
    \delta p_{hip}(\theta_{sk}, \theta_{sa}) = (l_{st} + l_{ss}) \theta_{sk} + l_{ss} \theta_{sa},
\end{equation*}
where $l_{ss},l_{st}$ are the length of the stance shin and stance thigh, and $\theta_{sk}, \theta_{sa}$ are the stance knee and stance ankle joint angles.
Previous research showed this value to approximately linearly increase during a human step \cite{jiang2012outputs}. The phase variable is defined as:
\begin{equation}\label{eq:phase_variable}
    \tau(q) = \frac{\delta p_{hip}(\theta_{sk}, \theta_{sa}) - \delta \underline{p}_{hip}}{\delta \overline{p}_{hip} - \delta \underline{p}_{hip}},
\end{equation}
where $\delta \underline{p}_{hip}$ and $\delta \overline{p}_{hip}$ are the initial and final hip positions in a step, respectively. To specify the walking behavior, we define a set of outputs for each domain, parameterized by parameters $\alpha_v := \{\alpha_{v, k}\}_{k =  1 \dots n_{y,v}}$, where $\alpha_{v, k}$ is the parameter set for the $k$-th output in $\Domain_v$, $n_{y,v}$ is the number of outputs for $\Domain_v$. In particular, state-based control for walking requires a velocity-modulating output $y_{1, v}(q, \dot{q})$ to progress the trajectory forward:
\begin{equation*}
    y_{1, v}(q, \dot{q}) = y_{1, v}^a(q, \dot{q}) - y_{1, v}^d(\alpha_v) \in \mathbb{R},
\end{equation*}
where $y_{1, v}^a(q, \dot{q}) = \dot{\delta} p_{hip}(q)$, the forward hip velocity, and $y_{1, v}^d(\alpha_v) \equiv v_{hip, v}$, a constant determined through optimization to match the constant hip velocity observed in experiment. To track the joint trajectories, we define 5 relative degree 2 outputs:
\begin{equation*}
    y_{2, v}(q) = y_{2, v}^a(q) - y_{2, v}^d(\tau(q), \alpha_v) \in \mathbb{R}^5.
\end{equation*}
Here $y_{2, v}^a(q)$ are all of the individual joint angles except the stance ankle.
Therefore, $n_{y,v}=6$ for the single support domains since $y_{2,v}\in\mathbb{R}^5$ and $y_{1,v}\in\mathbb{R}$.

\newsec{Joint Trajectories from Motion Capture Data.}
The parameterization of the outputs are via  
\Bezier curves. A \Bezier curve is a parameterized polynomial of variable $s\in[0,1]$ as
\begin{equation*}
    \mathbf{B}(s)=\sum\nolimits_{i=0}^{m}\alpha_i\frac{m!}{(m-i)!i!}s^i(1-s)^{m-i},
\end{equation*}
where $m$ is the degree of the \Bezier curve and $\left\{\alpha_i\right\}$ are the \Bezier coefficients. This provides a convenient way to parameterize nonlinear curves because simple manipulations of the \Bezier coefficients can give the derivative, integral, and square of the \Bezier curve. 

\begin{figure}
\label{fig:bez_cast_frost} 
\centering
\includegraphics[width=1\columnwidth]{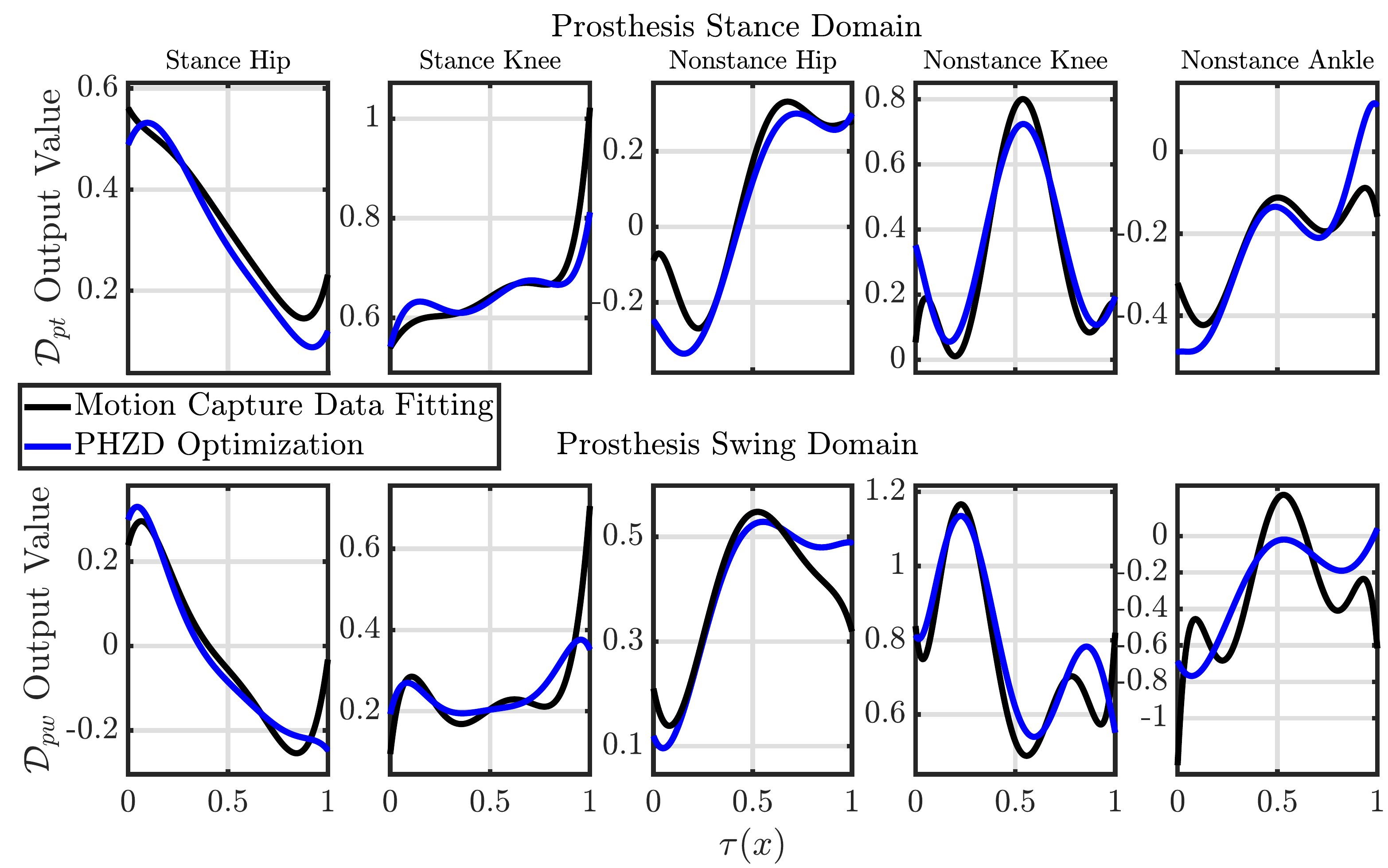}
\vspace{-0.7cm}
{\caption{The \Bezier polynomials for $y^d_{2, ps}(\tau(q), \alpha_{ps}$ (top) and $y^d_{2, pns}(\tau(q), \alpha_{pns})$: (black) motion capture data fit,  (blue) PHZD optimization result.}}
\vspace{-0.4cm}
\end{figure}
To divide the walking data into two phases, we examine $\delta p_{hip}(\theta_{pk}, \theta_{pa})$
which oscillates during walking.
We consider the portion when $\delta p_{hip}(\theta_{pk}, \theta_{pa})$ monotonically increases as the prosthesis stance phase, and the portion when $\delta p_{hip}(\theta_{pk}, \theta_{pa})$ monotonically decreases as the prosthesis non-stance phase. Based on this division we find $\delta \underline{p}_{hip}$ and $\delta \overline{p}_{hip}$ from the data for each step, used in \eqref{eq:phase_variable} to compute $\tau$. This gives us a sequence $\tau(1:T)$, where $T$ is the number of data time-steps for each walking step. For this range, the joint trajectories from Section \ref{ssec:ProcessMCData} are used to obtain the desired outputs $y^d_{v, k}$ , where the set $\{y^d_{v, k}\}_{k = 1 \dots n_{y, v}} = \{y^d_{1, v}, y^d_{2, v}\}$,
giving us a sequence $y^d_{v, k}(1:T)$.

We perform the curve fitting with a regression procedure since a \Bezier curve is a linear combination of the nonlinear basis functions $\mathbf{B}_i(s)=\frac{m!}{(m-i)!i!}s^i(1-s)^{m-i}$. 
Given the sequence of $\tau(1:T)$ and the corresponding sequence of output $y^d_{v, k}(1:T)$, let \begin{equation*}
    H\in\mathbb{R}^{T\times (m+1)},H_{ij}=\mathbf{B}_{j-1}(\tau(i))
\end{equation*}
be the regressor, then $\alpha^{mc}_{v, k} = (H^\intercal H)^{-1}(H^\intercal y^d_{ v, k}(1:T))$, where $\alpha^{mc}_{v, k}$ are the \Bezier coefficients fit to the motion capture data for $\Domain_v$ and output $k$. One special case is when $m=0$, in this case the Bezier regression is equivalent to taking the average, which is the used for $y_{1,v}^d$. Note that we do not require $\tau(1:T)$ to be unique or monotonic.

\newsec{Stable Walking Trajectory.}
To use the regression to get a state-based reference output from the data, the $\tau$ and $y^d_{v, k}$ sequences corresponding to multiple gaits are stacked and a set of \Bezier coefficients $\{\alpha^{mc}_{v, k}\}$ defining a single \Bezier curve is solved for each output for $k = 1,\dots,n_{y,v}$ and $v \in V$.
While we could track these outputs over the continuous domain, yielding invariant zero dynamics \cite{ZeroDynIsidori, westervelt2003HybridZero}, the zero dynamics may not remain invariant through impacts. 
Hence, we develop desired trajectories $y^d_v(\tau(q), \alpha_v)$ similar to those defined by $\alpha^{mc}_{v}$ that yield a stable walking gait, where $\alpha^{mc}_{v} := \{ \alpha^{mc}_{v, k} \}_{k = 1 \dots n_{y, v}}$.
Since the impact map causes a jump of velocities, we do not enforce an impact invariance condition on the velocity-modulating output, only the relative degree 2 outputs, rendering \emph{partial zero dynamics}:
\begin{equation*}
\small
\textbf{PZ}_{\alpha_v} = \{(q, \dot{q}) \in \mathcal{D}_v: \, y_{2, v}(q, \alpha_v) = 0,\, \dot{y}_{2, v}(q, \dot{q}, \alpha_v) = 0 \}.
\end{equation*}
We enforce \textit{partial hybrid zero dynamics} (PHZD) constraints \cite{ames2014human} while
minimizing the differences between the output defined by $\alpha_v$ and the motion capture outputs defined by $\alpha^{mc}_v$ with the following optimization:
\begin{align*} \tag{PHZD Optimization}
    c_v^\star = &\mathop {\arg\min }\limits_{{\alpha_v, \delta \overline{p}_{hip, v}, \delta \underline{p}_{hip, v}}} \; \mathcal{J}_v 
    \\ \tag{PHZD}
    &\quad\quad \mathrm{s.t} \quad \Delta_{e_i}(\Guard_{e_i} \cap \textbf{PZ}_{\alpha_{v_i}}) \subseteq \textbf{PZ}_{\alpha_{v_{i + 1}}},
\end{align*}
where $v_{i + 1}$ is the next domain in the directed cycle and 
\begin{equation*}
\resizebox{.95\hsize}{!}{$
    \begin{aligned}
    \mathcal{J}_v= \sum\nolimits_{k=1}^{n_{y, v}}\int_{0}^{1}w_{v, k}(y^d_{v, k}(s, \alpha_{v, k})-y^d_{v, k}(s, \alpha^{mc}_{v, k}) - \delta^{\alpha}_{v, k})^2 ds 
    \\
    + w_{\delta} \big((\delta \overline{p}_{hip, v}^{mc} - \delta \underline{p}_{hip, v}^{mc})
    -
    (\delta \overline{p}_{hip, v} - \delta \underline{p}_{hip, v}) \big)^2
    \end{aligned}
    $}
\end{equation*}
$\mathcal{J}_v$ can be represented as a simple quadratic function of the \Bezier coefficients and phase variable parameters. Here $\delta^{\alpha}_{v} := \{\delta^{\alpha}_{v, k} \}_{k = 2 \dots n_{y, v}}$ is a set of offsets the optimization can select to minimize the differences of the relative degree 2 outputs, since a joint offset likely existed in the data collection to determine $\alpha^{mc}_{v}$. We selected weights $w_{k, v}$ to encourage the optimization to give higher priority to outputs that were more difficult to match. 
We also include the difference of phase parameters to yield a trajectory with a similar step length, where $(\delta \overline{p}_{hip, v}^{mc}, \delta \underline{p}_{hip, v}^{mc})$ are the average phase parameters found for $\Domain_v$ in the motion capture data. 
The solution of the optimization is the set $c^\star_v = \{ \alpha_v^\star, \, \delta^{\alpha, \star}_{v},\, \delta \overline{p}^\star_{hip, v},\, \delta \underline{p}^\star_{hip, v} \}$ for each domain $\Domain_v$, where $\alpha^\star_v$, which includes $v_{hip}$, defines the desired output functions to render stable human-prosthesis walking similar to that seen in experiment. 
We solve this optimization in a direct collocation based multi-domain HZD gait optimization approach, called FROST, described in \cite{hereid20163d}.
Fig. 4 shows the comparison between the outputs $y^d_{2, v}(\tau(q), \alpha_v^\star)$ and $y^d_{2, v}(\tau(q), \alpha^{mc}_v) + \delta^{\alpha}_v$.

\subsection{Socket Force Estimation} \label{ssec:SocketForce}
Developing a model-based controller for the prosthesis requires knowledge of the interaction forces and moments between the human and prosthesis. This is at the socket for an amputee and at the pin connection between the iWalk adapter and top of prosthesis for our system. For simplicity, we refer to these forces and moments as the \textit{socket force}. This section outlines a method to estimate these forces offline based on motion capture data and the human-prosthesis model.

\newsec{Socket Force Profiles from Data Playback.}
To estimate the socket force present in the walking observed by motion capture, we simulate the human-prosthesis model following the joint trajectories from the data for each prosthesis stance phase.
The trajectory of each joint is fit with a \Bezier polynomial with parameters $\{\alpha_{v, i}^{pb}\}$, per the methods of Section \ref{ssec:GaitDesign}, for each data set $i$ of prosthesis stance. This process gives us 20 sets of joint trajectories to simulate. A feedback linearizing controller \cite{ames2014human} in simulation calculates the necessary torque $u$ at each joint to track these trajectories. 
By solving \eqref{eq:robotDyn} for $\ddot{q}$ and substituting into \eqref{eq:holoConstr} along with this $u$, we calculate the fixed joint constraint wrenches:
\begin{equation} \label{eq:Fsocket}
\vspace{-4pt}
    F_{s} = (J_{s} D^{-1} J_{s}^T)^{-1} (J_{s} D^{-1}(H - Bu) - \dot{J}_{s} \dot{q}),
\end{equation}
where $J_{s}$ is the Jacobian of the fixed joint holonomic constraint. Note we dropped the $v$ subscript since $J_{s}$ is the same for $\Domain_{ps}$ and $\Domain_{pns}$. We consider $F_s$ to be an approximation of the socket force seen by the prosthesis in the human-prosthesis walking experiment.

\newsec{Socket Force Tube.}
By calculating this socket force profile for 20 steps of walking data, we obtain a collection of 20 force profiles around which we define a tube. First we remove the socket force segment at the beginning and end since these sections correspond to a double support phase, which is outside the scope of this paper.
We use the $x$ coordinate of the markers on the human foot to determine whether the human is on single support by the prosthesis or double support, and remove the double support portion of the data. After this removal, the single support portion starts at $\underline{\tau}$ and ends at $\overline{\tau}$.
\begin{figure}[t]
\label{fig:force_tube} 
\centering
\includegraphics[width=1\columnwidth, trim={0 0 0 0},clip]{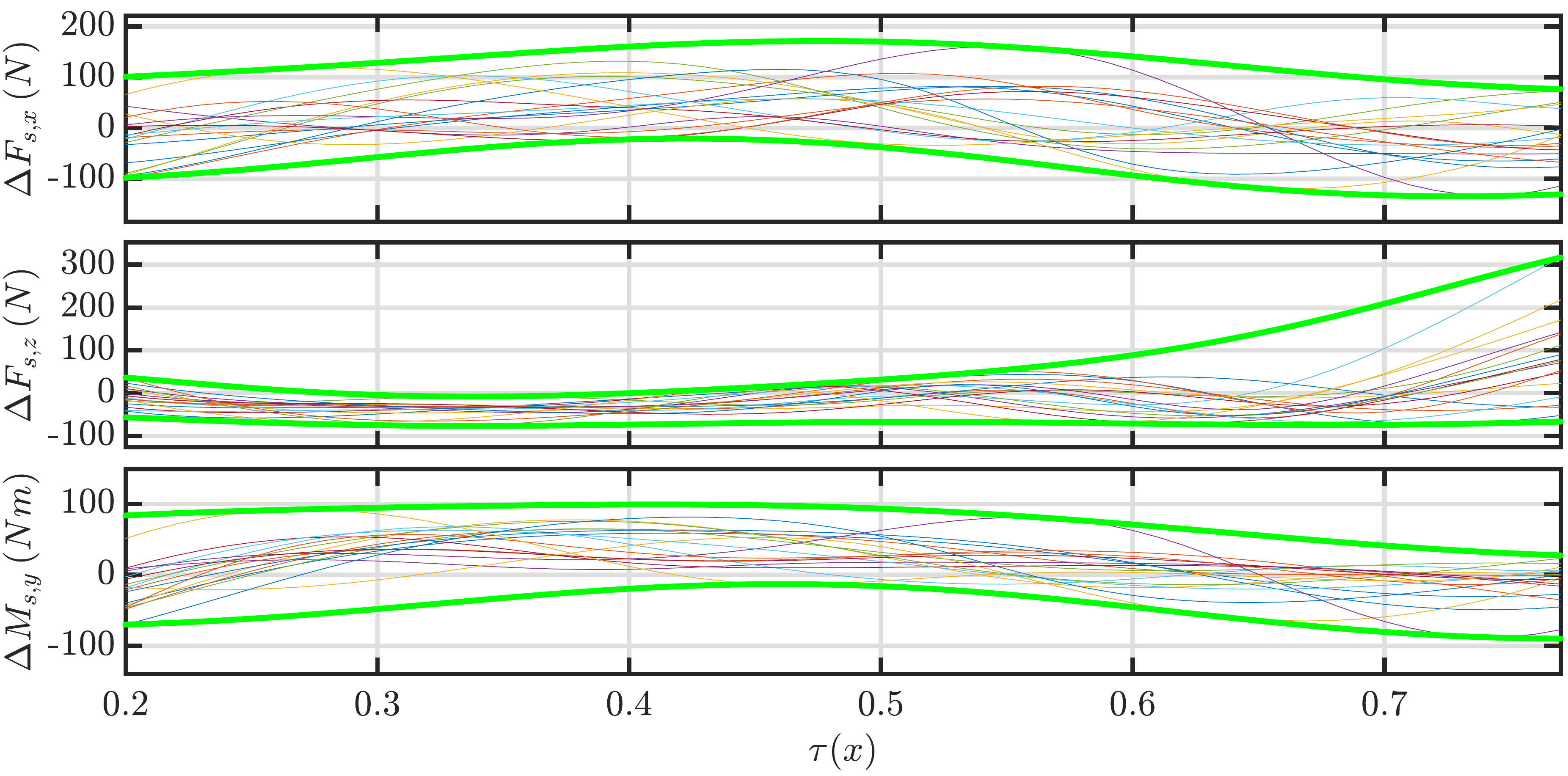}
\vspace{-0.5cm}
{\caption{Difference in socket force profile between nominal trajectory and data playback (colored lines). Computed force tube (green line).}}
\end{figure}
For the $N$ gaits collected from motion capture, let $F^i_s$ denote the socket force corresponding to the $i$-th gait computed. 
We compare these socket force profiles to $F^\star_s$ - the socket force profile from the nominal trajectory obtained by the FROST optimization. The difference between them is denoted as $\Delta F_s$, which includes 3 elements: $[\Delta F_{s,x}, \Delta F_{s,z}, \Delta M_{s,y}]$, denoting the longitudinal, vertical force and the pitch moment.
Then the following optimizations are used to find the upper bound $\Delta\overline{F_s}$ and lower bound $\Delta\underline{F_s}$ of $\Delta F_s$ as a function of $\tau$, shown in Fig. 5.
\begin{equation*}
\begin{aligned}
    \mathop{\max}\limits_{\Delta\underline{F_s}}& \int_{\underline{\tau}}^{\overline{\tau}}{\Delta\underline{F_s}(\tau)d\tau}\\
    s.t.\;&\forall i=1,...,N,\forall \overline{\tau}\ge \tau \ge \underline{\tau},\Delta\underline{ F_s}(\tau)\le F^i_s(\tau)-F^\star_s(\tau)\\
    \mathop{\min}\limits_{\Delta\overline{F_s}}& \int_{\underline{\tau}}^{\overline{\tau}}{\Delta\overline{F_s}(\tau)d\tau}\\
    s.t.\;&\forall i=1,...,N,\forall \overline{\tau}\ge \tau \ge\underline{\tau},\Delta\overline{ F_s}(\tau) \ge F^i_s(\tau)-F^\star_s(\tau).\\
    \end{aligned}
\end{equation*}
The upper and lower bounds $\Delta\overline{F_s}$ and $\Delta\underline{F_s}$ are represented as \Bezier curves of $\tau$, and the integration is computed as a linear function of the \Bezier coefficients.
\begin{rem}
Since the tube is computed with a finite set of measurements, the credibility of the force tube can be analyzed with the theory of Random Convex Programs \cite{calafiore2010random}. In general, using a high degree \Bezier curve results in a tighter tube, yet it hurts the credibility, i.e., there is a higher chance that the tube is breached by additional measurements.
\end{rem}

\section{ROBUST CLF-QP CONTROLLER} \label{sec:RobustCLF}
In this section we present the robust CLF-QP controller for the prosthesis in stance.

\newsec{Prosthesis Model and Dynamics.}
Since no sensing and actuation is available on the human body we model the prosthesis as an independent robotic leg, Fig. 3, per the methods of \cite{gehlhar2019control}, with 2 joints being the ankle and the knee and 3 limbs being the foot, shin, and thigh. For this subsystem, a floating base frame at the top of the prosthesis is subject to external forcing of the socket force $F_s$ present in the full model. A holonomic constraint enforces the foot to stay flat on the ground during prosthesis stance phase. The dynamics are given by:
\begin{align*} 
    \bar{D}(\bar{q}) \ddot{\bar{q}} + \bar{H}(\bar{q}, \dot{\bar{q}}) &= \bar{B} \bar{u} + \bar{J}^T(\bar{q}) \bar{F}(\bar{q}, \dot{\bar{q}}) + \bar{J}_{s}^T(\bar{q}) F_s,
    \\ 
    \dot{\bar{J}} \dot{\bar{q}} + \bar{J} \ddot{\bar{q}} &= 0,
\end{align*}
where $\bar{q} = (\bar{\q}_B^\intercal, \, q_p^\intercal)^\intercal$ and $\bar{\q}_B \in SE(2)$ represent the position and rotation of the subsystem's base frame $\bar{\mathcal{R}}_B$ with respect to the world frame $\mathcal{R}_W$. Again $q_p = (\theta_{pk},\, \theta_{pa})^\intercal$.
Here $\bar{J}(q)$ is the Jacobian of the foot's holonomic constraint and $\bar{J}_s$ is the projection of $F_s$ onto the base coordinates, see \cite{gehlhar2019control} for details.
Because of the holonomic constraint, the dynamics are written as a 4 state system:
\begin{equation*}
    {{\dot x}_p} = {f_p}({x_p}) + g_p^m({x_p}){u_p} + g_p^s({x_p}){F_s},
\end{equation*}
where $x_p = (\theta_{pk},\,  \theta_{pa},\, \dot{\theta}_{pk},\, \theta_{pa})^\intercal$ and $u_p \in \mathbb{R}^2$ denotes the motor torque input at the prosthesis knee and ankle, and $F_s$ denotes the socket force.

\newsec{Prosthesis Robust CLF-QP.}
The outputs for the prosthesis are defined as a subset of the outputs for the full system: $\bar{y}^a_1(q_p, \dot{q}_p) = \delta \dot{p}_{hip}(\theta_{pk}, \theta_{pa})$ and $\bar{y}^a_2(q_p) =  \theta_{pk}$.
With the output defined, let 
\begin{equation*}
\begin{aligned}
\eta_1 &=\bar{y}^a_1(q_p, \dot{q_p})-v_{hip}^\star &\doteq \bar{y}_1(x_p, v_{hip}^\star)\\
    \eta_2&=\bar{y}^a_2(q_p)-\bar{y}^d_2(\tau(q_p), \alpha_{ps}^\star)&\doteq \bar{y}_2(x_p, \alpha_{ps}^\star),
    \end{aligned}
\end{equation*}
where $v_{hip}^\star$ is the nominal hip velocity and $\alpha_{ps}^\star$ is the \Bezier coefficients corresponding to the nominal output trajectories, both determined through optimization for the full system $\Domain_{ps}$.
The output dynamics are obtained with feedback linearization (see \cite{ames2012control} for detailed derivation):
\begin{equation} \label{eq:output_dyn}
    \underbrace{
    \begin{bmatrix} \dot{\eta}_1\\\dot{\eta}_2\\ \ddot{\eta}_2\end{bmatrix}
    }_\eta
    =
    \underbrace{
    \begin{bmatrix}
    0&0&0\\0&0&1\\0&0&0
    \end{bmatrix}
    }_A
    \begin{bmatrix} \eta_1\\ \eta_2\\ \dot{\eta}_2\end{bmatrix} + 
    \underbrace{
    \begin{bmatrix} 1&0 \\ 0&0 \\0&1 \end{bmatrix}
    }_B
    \nu(u_p, F_s),
\end{equation}
where:
\begin{equation} \label{eq:nu}
    \begin{aligned}
   \nu(u_p, F_s)
    =
    \underbrace{
    \begin{bmatrix}
    \mathcal{L}_{f_p} \bar{y}_1 \\
    \mathcal{L}^2_{f_p} \bar{y}_2
    \end{bmatrix}
    }_{\mathcal{L}_{f_p}^*(x_p)}
    +
    \underbrace{
    \begin{bmatrix}
    \mathcal{L}_{g_p^m} \bar{y}_1 \\
    \mathcal{L}_{g_p^m}\mathcal{L}_{f_p} \bar{y}_2
    \end{bmatrix}
    }_{\mathcal{A}_m^*(x_p)}
    u_p
    +
    \underbrace{
    \begin{bmatrix}
    \mathcal{L}_{g_p^s} \bar{y}_1 \\
    \mathcal{L}_{g_p^s}\mathcal{L}_{f_p} \bar{y}_2 
    \end{bmatrix}
    }_{\mathcal{A}_s^*(x_p)}
    F_s
    \end{aligned}
\end{equation}
Here $\mathcal{L}_{f_p},\, \mathcal{L}_{g_p^m},\, \mathcal{L}_{g_p^s},\, \mathcal{L}_{f_p}^2$ denote the Lie derivatives \cite{IsidoriNonlinSyst}.
With $0< \varepsilon < 1$, the following rapidly exponentially stabilizing CLF \cite{ames2014rapidly} is defined:
\begin{equation*}
    V_{\varepsilon}(\eta) = 
    \eta 
    \begin{bmatrix}
    \frac{1}{\varepsilon} I & 0 \\
    0 & I
    \end{bmatrix}
    P
    \begin{bmatrix}
    \frac{1}{\varepsilon} I & 0 \\
    0 & I
    \end{bmatrix}
    :=\eta^\intercal P_\varepsilon \eta,
\end{equation*}
where $P$ is obtained by solving a Riccati equation with the linear output dynamics in \eqref{eq:output_dyn} with $Q$ and $R$ matrices representing the state and input costs:
\begin{equation*}
    {A^\intercal}P + PA - PB{R^{ - 1}}{B^\intercal}P + Q = 0.
\end{equation*}

Defining a CLF-QP for the prosthesis would include $\nu(u_p, F_s)$ from \eqref{eq:nu}, requiring knowledge of the socket force $F_s$. Since this is unknown, we instead use an estimate of the range of $F_s$ obtained from the analysis in Section \ref{ssec:SocketForce}:
\[
F_s^\star(\tau)+\Delta \overline{F_s}(\tau)\ge F_s(\tau)\ge F_s^\star(\tau)+\Delta \underline{F_s}(\tau),
\]
where $\tau(q_p)$ is the phase variable. Then the following robust CLF-QP is formulated that enforces the CLF condition on all possible $F_s$ within the range:
\begin{equation}\label{eq:robust_CLF_QP}
\resizebox{.9\hsize}{!}{$
\begin{aligned}
u_p^ \star = \mathop {\arg\min }\limits_{{u_p \in \mathbb{R}^2}} \,\, &u_p^\intercal \mathcal{H} u_p + b^\intercal u_p
\\
\textrm{s.t.} \,\, &\forall F_s^\star(\tau)+\Delta \overline{F_s}(\tau)\ge F_s\ge F_s^\star(\tau)+\Delta \underline{F_s}(\tau),
\\
& \mathcal{L}_B V_\varepsilon(\eta) \nu(u_p, F_s) 
\le 
-\frac{\gamma}{\varepsilon} V_{\varepsilon} - \mathcal{L}_A V_\varepsilon(\eta),
\end{aligned}
$}
\end{equation}
which is a QP w.r.t. $u_p$. Here $u_p$ is a function of $x_p,\, c_v^*,\, F_s^*,\, \Delta \overline{F_s},\ \textrm{and } \Delta \underline{F_s}$, $\gamma > 0$, 
\begin{equation*}
    \begin{aligned}
    \mathcal{H} &= \mathcal{A}_m^*(x_p)^\intercal \mathcal{A}_m^*(x_p), 
    \\
    b &= (\mathcal{L}_{f_p}^*y(x_p) + \mathcal{A}_s^*(x_p) F_s^*)^\intercal \mathcal{A}_m^*(x_p),
    \\
    \mathcal{L}_A V_\varepsilon(\eta) &= \eta ^T(P_{\varepsilon}A + {A^T}P_{\varepsilon})\eta, \quad \mathrm{and}
    \\
    \mathcal{L}_B V_\varepsilon(\eta) &= 2 {\eta ^T}P_{\varepsilon}B.
    \end{aligned}
\end{equation*}
The CLF condition enforces $\eta$ to converge exponentially to the origin, thus driving the system to track the desired trajectory \cite{ames2014rapidly}.
When $\Delta_{F_s}(\tau)$ is a hyperbox, \eqref{eq:robust_CLF_QP} is easily solved as a QP.

The formulation of this controller can yield a non-smooth control input since it stabilizes for the worst-case scenario in a point-wise optimal way, and the worst case changes at each point. To smooth the control input profile to be physically feasible on the prosthesis platform, we modify the QP: 
\begin{align}\label{eq:robust_CLF_QP_relax}
\mathop {\arg\min }\limits_{{(u_p, \rho) \in \mathbb{R}^3}} \,\, &u_p^\intercal \mathcal{H} u_p + b^\intercal u_p + c_\rho \rho
\\
\textrm{s.t.}\,\,  &\forall F_s^\star(\tau)+\Delta \overline{F_s}(\tau)\ge F_s\ge F_s^\star(\tau)+\Delta \underline{F_s}(\tau),
\notag\\
& \mathcal{L}_B V_\varepsilon(\eta) \nu(u_p, F_s) 
\le 
\frac{\gamma}{\varepsilon}(\rho - V_{\varepsilon}) - \mathcal{L}_A V_\varepsilon(\eta),
\notag
\\ 
& -\Delta_u \le (u_p - \bar{u}_p) \leq \Delta_u, \notag
\\
& 0 < \rho < \rho_{max},
\notag
\end{align}
where the bound $\Delta_u$ is placed on the change between $u_p$ and the previous control input $\bar{u}_p$ to minimize the fluctuations in $u_p$. To ensure the QP is always solvable with the torque bounds, the relaxation term $\rho$ is added to the Lyapunov function $V_\varepsilon$. With upper bound $\rho_{max}$, this relaxation term allows the QP to find a solution within a Lyapunov level set less than $\rho$, which is penalized with the weight $c_\rho$.

\begin{figure}
\centering
\includegraphics[width=1\columnwidth, trim={0 0 0 0},clip]{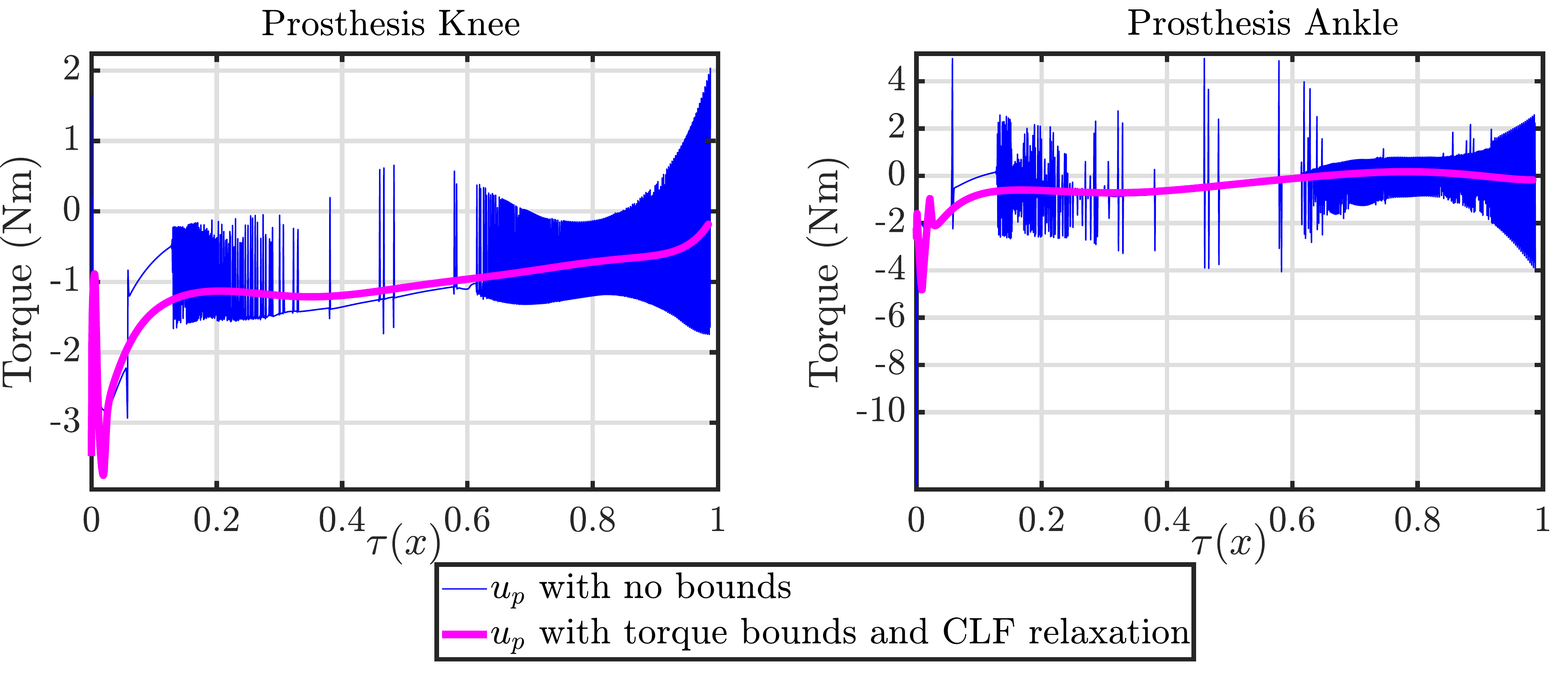}
\caption{Prosthesis control inputs calculated with \eqref{eq:robust_CLF_QP} (blue) with no torque bounds and \eqref{eq:robust_CLF_QP_relax} (pink) with torque bounds and CLF relaxation for smoothing.}
\vspace{-0.6cm}
\end{figure}

\newsec{Results.}
In simulation the human side tracked the nominal trajectory $y_{2, v}^d(\tau(q), \alpha_v^\star)$ from Section \ref{ssec:GaitDesign} with a feedback linearizing controller. The prosthetic tracked this trajectory with the robust CLF-QP controller in $\Domain_{ps}$ and feedback linearization in $\Domain_{pns}$. In $\Domain_{ps}$, the controller presented in \eqref{eq:robust_CLF_QP} was used and the control input exhibited rapid chatter, as shown in Fig. 6. Hence, the robust CLF-QP was modified as \eqref{eq:robust_CLF_QP_relax} to reduce the change in $u_p$ through the use of torque bounds and a CLF relaxation. Fig. 6 shows this smoothed torque profile.
The phase portraits of 20 steps shown in Fig. 7 show this novel prosthesis controller \eqref{eq:robust_CLF_QP_relax} achieves stability while accounting for a range of force disturbances. Fig. 7 also shows the phase portraits of the motion capture data for individual steps. Its alignment with the simulation phase portraits shows our nominal trajectory from optimization represents the walking observed by motion capture well. Note that the jump that appears in the simulation portrait but not in the motion capture data is a result of the rigid impact model we use for the human-prosthesis system. A human has more compliance that absorbs some impact yielding data that does not show a large discrete jump in velocities. The alignment between the portraits in combination with the stability shown support the idea that our controller would achieve stability in a similar human-prosthesis walking experiment.

To test this controller's robustness to perturbations in the nominal force profile, we simulated the human-prosthesis system with the human joints following a different trajectory than the nominal trajectory with feedback linearization and gave the prosthesis an initial condition off of its nominal trajectory. To enforce the nominal prosthesis trajectory, we tested 3 different controllers on the prosthesis in $\Domain_{ps}$: our robust CLF-QP \eqref{eq:robust_CLF_QP_relax}, the robust-passive controller we used for the motion capture experiments \cite{azimi2017robust}, and a PD controller. The results for 3 different human gaits in Fig. 8 demonstrate the benefit of model-dependent controllers since the robust passive controller with some model dependence outperforms the PD controller in tracking performance with the perturbation. Further, our robust CLF-QP outperformed both of these controllers with its consideration of the human-prosthesis interaction forces. Even more significant than the outperformance in tracking is that the robust-passive controller required careful tuning of the PD gains for the ankle and did not successfully operate before $\tau(x) = 0.2$ for the perturbed trajectories. Here the prosthesis would fall backwards instead of progressing forward along the trajectory, whereas our robust CLF-QP performed successfully for the whole range without gain tuning. These results also demonstrate the success of our controller design: it stabilized trajectories with force profiles different from the expected nominal force profile by considering a set of forces and stabilizing for the worst case scenario. This robustness is imperative for a prosthesis connected to a human with varying behavior and where stability is essential for the human's safety.

\begin{figure} [t]
\centering
\includegraphics[width=1\columnwidth]{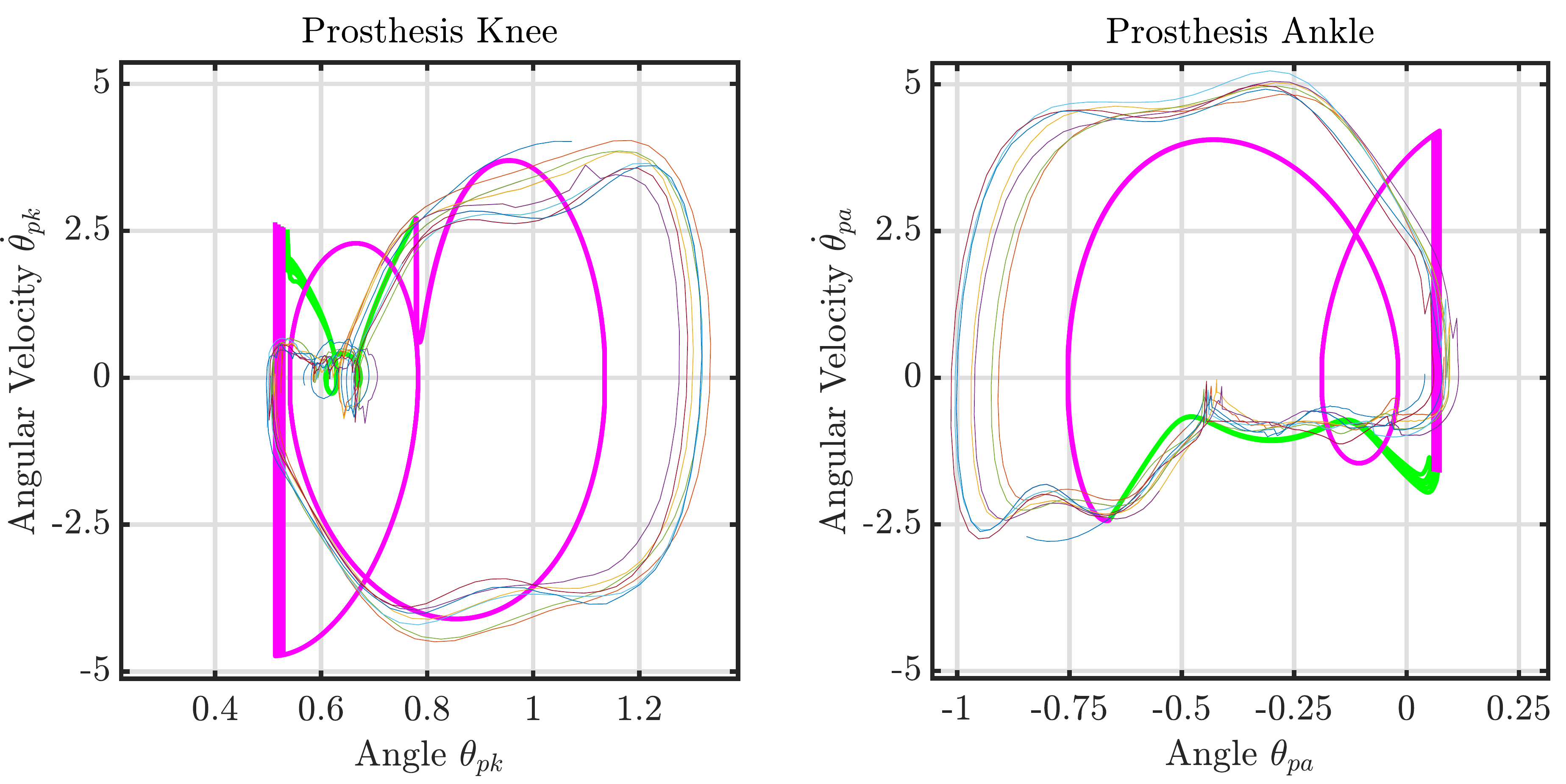}
{\caption{Phase portraits: prosthesis knee and ankle with robust CLF-QP controller in $\Domain_{ps}$ (green) and feedback linearization in $\Domain_{pns}$ (pink),  (multi-colored lines) motion capture data for individual steps.}}
\end{figure}

\begin{figure}
\centering
\includegraphics[width=1\columnwidth, trim={0 0 0 0},clip]{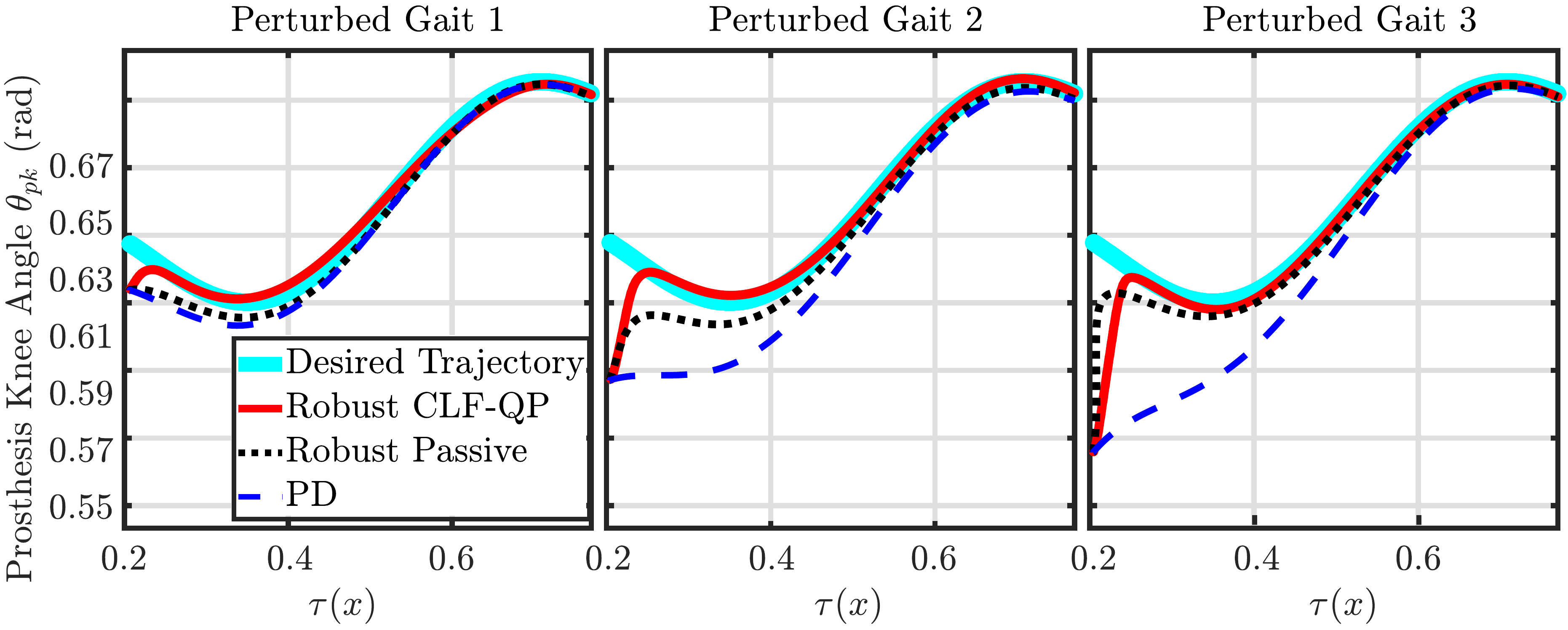}
\caption{Prosthesis knee joint angle trajectories from $\Domain_{ps}$ from simulations with 3 controllers with 3 perturbed human trajectories, (light blue) desired trajectory.}
\label{fig:robust_test}
\vspace{-0.4cm}
\end{figure}

\section{CONCLUSION AND FUTURE WORK} \label{sec:Conclusion}
This paper presents a methodology that models the human walking behavior while wearing a powered prosthesis and uses the model information to design a model-dependent controller for the prosthesis. 
The data obtained from motion capture is processed and used in two ways. First, a customized gait optimization procedure is proposed to extract the nominal trajectory from the data, which emulates the human walking recorded by motion capture, and satisfies the dynamics of the human-prosthesis system. Second, a ``playback'' procedure is designed to obtain the interaction force profiles from the multiple steps recorded, which are then used to construct a force tube that contains all the force profiles. With this information, a robust CLF-QP controller is designed that guarantees convergence to the nominal trajectory.
Simulation results demonstrate the robustness of this prosthesis controller compared to a model independent controller as well prosthesis controllers with some model dependence.
This novel methodology of characterizing human interaction in prosthesis walking provides a means to replace the need for an expensive force sensor with a single set of motion capture experiments for a given user, preventing the introduction of noise from the force sensor while also increasing the robustness of a model-dependent CLF-QP.

For future work, this controller can be experimentally realized on the prosthesis platform to assess its tracking performance, energy efficiency, and robustness to disturbances compared to the current PD controller and robust passive controller used for trajectory tracking. The advantage of this method can be tested across multiple subjects by having each subject be part of one set of motion capture experiments and then develop a specific trajectory and force tube based on their model parameters and data. The same user can test their specific controller in experiment. Examining the results across multiple subjects could also provide insight on the generalizability of force profiles across users such that the force profile could be predicted for a subject without motion capture.
In the case a force sensor is incorporated into the prosthesis platform to directly measure the interaction force, the force profiles collected in this work shall provide the statistics (mean and covariance) for the design of the force sensor filter, e.g., a Kalman filter. This methodology opens the door to model-dependent prosthesis controllers that account for the human's varying dynamic behavior and establish stability in response.

\bibliographystyle{IEEEtranS}
\bibliography{bibliography}

\begin{thebibliography}{10}
\providecommand{\url}[1]{#1}
\csname url@samestyle\endcsname
\providecommand{\newblock}{\relax}
\providecommand{\bibinfo}[2]{#2}
\providecommand{\BIBentrySTDinterwordspacing}{\spaceskip=0pt\relax}
\providecommand{\BIBentryALTinterwordstretchfactor}{4}
\providecommand{\BIBentryALTinterwordspacing}{\spaceskip=\fontdimen2\font plus
\BIBentryALTinterwordstretchfactor\fontdimen3\font minus
  \fontdimen4\font\relax}
\providecommand{\BIBforeignlanguage}[2]{{%
\expandafter\ifx\csname l@#1\endcsname\relax
\typeout{** WARNING: IEEEtranS.bst: No hyphenation pattern has been}%
\typeout{** loaded for the language `#1'. Using the pattern for}%
\typeout{** the default language instead.}%
\else
\language=\csname l@#1\endcsname
\fi
#2}}
\providecommand{\BIBdecl}{\relax}
\BIBdecl

\bibitem{ames2013human}
A.~D. Ames, ``Human-inspired control of bipedal robots via control {L}yapunov
  functions and quadratic programs,'' in \emph{Proceedings of the 16th
  international conference on Hybrid systems: computation and control}.\hskip
  1em plus 0.5em minus 0.4em\relax ACM, 2013, pp. 31--32.

\bibitem{ames2014human}
------, ``Human-inspired control of bipedal walking robots,'' \emph{IEEE
  Transactions on Automatic Control}, vol.~59, no.~5, pp. 1115--1130, 2014.

\bibitem{ames2012control}
A.~D. Ames, K.~Galloway, and J.~W. Grizzle, ``Control lyapunov functions and
  hybrid zero dynamics,'' in \emph{2012 IEEE 51st IEEE Conference on Decision
  and Control (CDC)}.\hskip 1em plus 0.5em minus 0.4em\relax IEEE, 2012, pp.
  6837--6842.

\bibitem{ames2014rapidly}
A.~D. Ames, K.~Galloway, K.~Sreenath, and J.~W. Grizzle, ``Rapidly
  exponentially stabilizing control {L}yapunov functions and hybrid zero
  dynamics,'' \emph{IEEE Transactions on Automatic Control}, vol.~59, no.~4,
  pp. 876--891, 2014.

\bibitem{PowAnkleFootStair}
S.~Au, M.~Berniker, and H.~Herr, ``Powered ankle-foot prosthesis to assist
  level-ground and stair-descent gaits,'' \emph{Neural Networks}, vol.~21,
  no.~4, pp. 654 -- 666, 2008, robotics and Neuroscience.

\bibitem{azimi2017robust}
V.~Azimi, T.~Shu, H.~Zhao, E.~Ambrose, A.~D. Ames, and D.~Simon, ``Robust
  control of a powered transfemoral prosthesis device with experimental
  verification,'' in \emph{American Control Conference (ACC), 2017}.\hskip 1em
  plus 0.5em minus 0.4em\relax IEEE, 2017, pp. 517--522.

\bibitem{calafiore2010random}
G.~C. Calafiore, ``Random convex programs,'' \emph{SIAM Journal on
  Optimization}, vol.~20, no.~6, pp. 3427--3464, 2010.

\bibitem{dasgupta1999making}
A.~Dasgupta and Y.~Nakamura, ``Making feasible walking motion of humanoid
  robots from human motion capture data,'' in \emph{Proceedings 1999 IEEE
  International Conference on Robotics and Automation (Cat. No. 99CH36288C)},
  vol.~2.\hskip 1em plus 0.5em minus 0.4em\relax IEEE, 1999, pp. 1044--1049.

\bibitem{HumanInertia}
W.~Erdmann, ``Geometry and inertia of the human body - review of research,''
  vol.~1, no.~1, pp. 23--35, 1999.

\bibitem{gehlhar2019control}
R.~Gehlhar, J.~Reher, and A.~D. Ames, ``Control of separable subsystems with
  application to prostheses,'' \emph{arXiv preprint arXiv:1909.03102v1}, 2019.

\bibitem{VirtConsCtrlProst}
R.~D. Gregg, T.~Lenzi, L.~J. Hargrove, and J.~W. Sensinger, ``Virtual
  constraint control of a powered prosthetic leg: From simulation to
  experiments with transfemoral amputees,'' \emph{IEEE Transactions on
  Robotics}, vol.~30, no.~6, pp. 1455--1471, Dec 2014.

\bibitem{ModelsGrizzle}
J.~W. Grizzle, C.~Chevallereau, R.~W. Sinnet, and A.~D. Ames, ``Models,
  feedback control, and open problems of 3d bipedal robotic walking,''
  \emph{Automatica}, vol.~50, no.~8, pp. 1955 -- 1988, 2014.

\bibitem{hereid20163d}
A.~Hereid, E.~A. Cousineau, C.~M. Hubicki, and A.~D. Ames, ``{3D} dynamic
  walking with underactuated humanoid robots: A direct collocation framework
  for optimizing hybrid zero dynamics,'' in \emph{Robotics and Automation
  (ICRA), 2016 IEEE International Conference on}.\hskip 1em plus 0.5em minus
  0.4em\relax IEEE, 2016, pp. 1447--1454.

\bibitem{CtrlWearRobot}
M.~A. {Holgate}, T.~G. {Sugar}, and A.~W. {Bohler}, ``A novel control algorithm
  for wearable robotics using phase plane invariants,'' in \emph{2009 IEEE
  International Conference on Robotics and Automation}, May 2009, pp.
  3845--3850.

\bibitem{ZeroDynIsidori}
A.~Isidori and C.~H. Moog, ``On the nonlinear equivalent of the notion of
  transmission zeros,'' in \emph{Modelling and Adaptive Control}, C.~I. Byrnes
  and A.~B. Kurzhanski, Eds.\hskip 1em plus 0.5em minus 0.4em\relax Berlin,
  Heidelberg: Springer Berlin Heidelberg, 1988, pp. 146--158.

\bibitem{IsidoriNonlinSyst}
A.~Isidori, \emph{Nonlinear Control Systems}, 3rd~ed., M.~Thoma, E.~D. Sontag,
  B.~W. Dickinson, A.~Fettweis, J.~L. Massey, and J.~W. Modestino, Eds.\hskip
  1em plus 0.5em minus 0.4em\relax Berlin, Heidelberg: Springer-Verlag, 1995.

\bibitem{jiang2012outputs}
S.~Jiang, S.~Partrick, H.~Zhao, and A.~D. Ames, ``Outputs of human walking for
  bipedal robotic controller design,'' in \emph{American Control Conference
  (ACC), 2012}.\hskip 1em plus 0.5em minus 0.4em\relax IEEE, 2012, pp.
  4843--4848.

\bibitem{lawson2012preliminary}
B.~E. Lawson, A.~Huff, and M.~Goldfarb, ``A preliminary investigation of
  powered prostheses for improved walking biomechanics in bilateral
  transfemoral amputees,'' in \emph{2012 Annual International Conference of the
  IEEE Engineering in Medicine and Biology Society}.\hskip 1em plus 0.5em minus
  0.4em\relax IEEE, 2012, pp. 4164--4167.

\bibitem{HybInvStabFeedLin}
A.~E. Martin and R.~D. Gregg, ``Hybrid invariance and stability of a feedback
  linearizing controller for powered prostheses,'' in \emph{2015 American
  Control Conference (ACC)}, July 2015, pp. 4670--4676.

\bibitem{MLS}
R.~M. Murray, S.~S. Sastry, and L.~Zexiang, \emph{A Mathematical Introduction
  to Robotic Manipulation}, 1st~ed.\hskip 1em plus 0.5em minus 0.4em\relax Boca
  Raton, FL, USA: CRC Press, Inc., 1994.

\bibitem{HumanParam}
S.~Plagenhoef, F.~G. Evans, and T.~Abdelnour, ``Anatomical data for analyzing
  human motion,'' \emph{Research Quarterly for Exercise and Sport}, vol.~54,
  no.~2, pp. 169--178, 1983.

\bibitem{Stein1987StancePC}
J.~L. Stein and W.~C. Flowers, ``Stance phase control of above-knee prostheses:
  knee control versus sach foot design.'' \emph{Journal of biomechanics}, vol.
  20 1, pp. 19--28, 1987.

\bibitem{DesignControlTransProsth}
F.~Sup, A.~Bohara, and M.~Goldfarb, ``Design and control of a powered
  transfemoral prosthesis,'' \emph{The International Journal of Robotics
  Research}, vol.~27, no.~2, pp. 263--273, 2008, pMID: 19898683.

\bibitem{wehner2009internal}
T.~Wehner, L.~Claes, and U.~Simon, ``Internal loads in the human tibia during
  gait,'' \emph{Clinical Biomechanics}, vol.~24, no.~3, pp. 299--302, 2009.

\bibitem{westervelt2003HybridZero}
E.~R. {Westervelt}, J.~W. {Grizzle}, and D.~E. {Koditschek}, ``Hybrid zero
  dynamics of planar biped walkers,'' \emph{IEEE Transactions on Automatic
  Control}, vol.~48, no.~1, pp. 42--56, Jan 2003.

\bibitem{westervelt2018feedback}
E.~R. Westervelt, J.~W. Grizzle, C.~Chevallereau, J.~H. Choi, and B.~Morris,
  \emph{Feedback control of dynamic bipedal robot locomotion}.\hskip 1em plus
  0.5em minus 0.4em\relax CRC press, 2018.

\bibitem{winter1991biomechanics}
D.~Winter, \emph{The Biomechanics and Motor Control of Human Gait: Normal,
  Elderly and Pathological}.\hskip 1em plus 0.5em minus 0.4em\relax University
  of Waterloo Press, 1991.

\bibitem{yang2015mechanical}
U.-J. Yang and J.-Y. Kim, ``Mechanical design of powered prosthetic leg and
  walking pattern generation based on motion capture data,'' \emph{Advanced
  Robotics}, vol.~29, no.~16, pp. 1061--1079, 2015.

\bibitem{zhao2016multi}
H.~{Zhao}, J.~{Horn}, J.~{Reher}, V.~{Paredes}, and A.~D. {Ames},
  ``Multicontact locomotion on transfemoral prostheses via hybrid system models
  and optimization-based control,'' \emph{IEEE Transactions on Automation
  Science and Engineering}, vol.~13, no.~2, pp. 502--513, April 2016.

\bibitem{zhao2017preliminary}
H.~Zhao, E.~Ambrose, and A.~D. Ames, ``Preliminary results on energy efficient
  {3D} prosthetic walking with a powered compliant transfemoral prosthesis,''
  in \emph{Robotics and Automation (ICRA), 2017 IEEE International Conference
  on}.\hskip 1em plus 0.5em minus 0.4em\relax IEEE, 2017, pp. 1140--1147.

\bibitem{zhao2017first}
H.~Zhao, J.~Horn, J.~Reher, V.~Paredes, and A.~D. Ames, ``First steps toward
  translating robotic walking to prostheses: a nonlinear optimization based
  control approach,'' \emph{Autonomous Robots}, vol.~41, no.~3, pp. 725--742,
  2017.

\end{thebibliography}

\end{document}